\documentclass[10pt,twocolumn,letterpaper]{article}

\usepackage[pagenumbers]{wacv}            

\usepackage[accsupp]{axessibility}  


\usepackage{xcolor}
\usepackage{multirow}
\usepackage{colortbl}
\usepackage{tikz}
\usepackage{pgfplots}
\usepackage{makecell}
\pgfplotsset{compat=1.18}

\definecolor{wacvblue}{rgb}{0.21,0.49,0.74}
\usepackage[breaklinks,colorlinks,allcolors=wacvblue]{hyperref}

\def\wacvPaperID{2743} 
\def\confName{WACV}
\def\confYear{2026}

\title{Saliency-Guided DETR for Moment Retrieval and Highlight Detection}

\author{
    Aleksandr Gordeev\thanks{Equally contributed} \\
    {\tt\small asegordeev@gmail.com}
    \and
    Vladimir Dokholyan\footnotemark[1] \\
    {\tt\small doholyan.vs@phystech.edu}
    \and
    Irina Tolstykh \\
    {\tt\small irinakr4snova@gmail.com}
    \and
    Maksim Kuprashevich\thanks{Corresponding author}  \\
    {\tt\small mvkuprashevich@gmail.com}
    \\ \\
    SALUTEDEV LLC
}

\newcommand{\ModelName}{SG-DETR}

\begin{document}

\maketitle

\begin{abstract} 
\label{section} 
With the rapid growth of video content available, the ability to search for specific moments within videos using textual queries has become increasingly relevant. This is crucial in many scenarios, from surveillance cameras where it may be necessary to find specific events in extensive video streams to searching for exciting movie scenes. However, existing approaches for video Moment Retrieval and Highlight Detection often struggle to effectively align text and video features, limiting their performance. We argue that utilizing recent foundational video models designed for video-text alignment can overcome these limitations. We propose a novel architecture that utilizes such models to test this hypothesis. Combined with our novel Saliency-Guided Cross Attention mechanism and a hybrid DETR architecture, our approach provides significantly improved results. To further enhance our approach, we developed InterVid-MR — a large-scale, high-quality dataset specifically designed for a pretraining stage. Extensive experiments and comparisons with current state-of-the-art methods confirm the effectiveness of the approach, achieving 58.8 mAP on QVHighlights, 60.7 R@1 mIoU on Charades-STA, and 42.4 R@1 mIoU on TACoS. These results highlight the efficiency and scalability of the method for video-language tasks in both zero-shot and fine-tuning scenarios. The dataset and code will be publicly available. \footnote{https://gracikk-ds.github.io/sg-detr/}
\end{abstract}
\section{Introduction}
\label{section:introduction}

The task of searching for specific moments in a video based on a text query has always been in demand. With the rapid growth of video content, it has become increasingly relevant. For instance, in the entertainment industry, quickly locating scenes of interest like epic battles in movies or specific landmarks in travel videos can greatly enhance content accessibility and user engagement. In sports analytics, identifying key events can be very important for streaming platforms and spectators. Similarly, in security surveillance, rapid search of incidents like car accidents, crime cases, or safety violations in industrial settings is vital for timely response or investigations.

Despite this demand, existing solutions for video moment retrieval (MR), which aims to localize specific video intervals, and highlight detection (HD), which focuses on assigning a relevance score to each video sub-clip, exhibit limited performance and scalability, constraining their practical adoption. Most of the methods utilize frozen encoders, such as SlowFast \cite{SlowFast} and CLIP \cite{CLIP}, to extract features from videos and text queries in the QVHighlights \cite{QVHighlights} benchmark, but they often fail to align these features effectively. Some works, such as \cite{R2}, have attempted to address CLIP's limitations and enhance the quality of video features and have even achieved some performance improvement. However, recent advances in foundational video models have enabled the use of more powerful encoders for both text and video, paving the way for novel architectural approaches in the moment retrieval field. We propose a new model block that utilizes the advantages of the InternVideo2 \cite{internvideo2} aligned video-text encoder. Our approach starts with generating initial highlight scores termed as Local Saliency Scores by directly comparing text and video embeddings and then refining them by incorporating the global context of the entire video.

Inspired by the success of DETR in 2D object detection \cite{DETR}, the authors of the QVHighlights benchmark adapted its principles for 1D video moment localization task. The resulting model, Moment-DETR \cite{QVHighlights}, became a foundational development in the field, paving the way for further works \cite{QD_DETR, CG_DETR, BAM_DETR, TR_DETR}. However, other architecture variants have also been proposed to address this problem. For instance, the authors of \cite{UniVTG} avoid the one-to-one matching typical for DETR-based models, and instead tackle moment localization using standard detection heads with one-to-many matching. Inspired by advancements in 2D detection \cite{CO_DETR}, we combine both strategies and propose a hybrid architecture for detecting relevant intervals, which incorporates a convolutional detector head and a transformer decoder block with a one-to-one matching scheme. Additionally, we use an IoU scoring mechanism to improve the localization of video spans further.

Different parts of a video vary in their relevance to a text query. To focus the model on the most relevant content, we propose a Saliency-Guided Cross Attention (SGCA) mechanism for cross-modal interaction, where cross-attention scores are weighted by the Local Saliency Scores (\cref{subsection:local_saliency_scores}). By avoiding extra tokens or complex gating modules, SGCA provides a simpler implementation compared to prior approaches \cite{CG_DETR, TR_DETR}.

Another area of research focuses on the interrelationship between the tasks of moment retrieval and highlight detection. Recently, the authors of TR-DETR \cite{TR_DETR} emphasized the need to ensure interaction between the heads solving these two tasks. We build upon these ideas and further integrate the tasks of moment retrieval and highlight detection using Saliency Amplifier module.

Despite the high quality of the QVHighlights dataset's annotations, its limited size necessitates additional data for effective model pretraining. To address this, we introduce a novel data generation approach that utilizes large video retrieval datasets, such as InterVid-FLT-10m \cite{InternvideoV1}, and leverages the impressive capabilities of foundational text-video models to create high-quality training data. This method enables us to build a pretraining dataset called InterVid-MR, which contains 150k samples. The model, trained solely on the InterVid-MR dataset, demonstrates strong performance in zero-shot evaluations on the QVHighlights benchmark. Moreover, fine-tuned on QVHighlights data, the model achieves state-of-the-art results.

The contributions of the work can be summarized as:
\begin{itemize}
    \item We developed a new architecture for moment retrieval and highlight detection tasks, leveraging the latest advancements in video-text alignment and object detection.
    \item We proposed a novel data generation approach and built a pre-training dataset with 150k samples.
    \item We conducted extensive experiments and achieved state-of-the-art results across three benchmarks: QVHighlights, Charades-STA, and TACoS.
    \item We provide publicly available training and inference code, InterVid-MR dataset and data generation scripts.
\end{itemize}

\section{Related Works} 
\label{section:related_work}

\paragraph{Video Temporal Grounding}
Video temporal grounding refers to linking a textual query to its corresponding segments within a video. This process involves two core components: moment retrieval and highlight detection. 

Approaches for moment retrieval task are typically classified into proposal-based and proposal-free methods. Proposal-based methods utilize predefined proposals, such as sliding windows \cite{MR_sliding_windows_1, MR_sliding_windows_2} or temporal anchors \cite{MR_temporal_anchors_1}. These methods often require complex pre- and post-processing, with performance being heavily influenced by the quality of the generated proposals. In contrast, proposal-free methods \cite{QVHighlights, UniVTG, CG_DETR} leverage multimodal information and directly predict the start and end coordinates of the target moments, thus bypassing the need for explicit proposal generation.

Methods for the highlight detection task are typically divided into three categories based on the level of annotation: supervised \cite{Supervised_HD_1, Supervised_HD_2}, which requires precise and often costly annotations; weakly supervised \cite{weakly_Supervised_HD_1, weakly_Supervised_HD_2}, which learns to identify key moments using event labels; and unsupervised \cite{UNSupervised_HD_1, UNSupervised_HD_2, UNSupervised_HD_3}, which operates without labeled data.

Traditionally, moment retrieval and highlight detection tasks have been addressed independently. However, a recent study \cite{QVHighlights} introduced the QVHighlights dataset, which facilitates the simultaneous resolution of both tasks within a unified framework. In conjunction with the dataset, the authors proposed a novel approach based on a DETR-like architecture designed to jointly tackle the challenges of both moment retrieval and highlight detection.

Building on the ideas proposed in \cite{QVHighlights}, QD-DETR \cite{QD_DETR} enhanced video features by integrating textual information through Cross-Attention mechanisms. The subsequent study by \cite{CG_DETR} introduced an Adaptive Cross-Attention module that incorporates dummy tokens to redirect attention, thereby minimizing the representation of irrelevant video clips in response to the text query. BAM-DETR \cite{BAM_DETR} extended DETR architecture by introducing a boundary-based prediction approach, where the model predicts segment boundaries rather than segment centers and their width. In \cite{TR_DETR}, the authors explored the relationship between moment retrieval and highlight detection tasks and proposed a joint prediction framework that aligns both tasks.

Another line of research diverges from DETR-like architectures. UMT \cite{UMT} introduced a transformer-based model that leverages multimodal data, integrating both video and audio streams to enhance performance. The most recent work,  Mr.BLIP \cite{MrBLIP}, proposed an approach based on LLMs, focusing exclusively on solving moment retrieval tasks.

\paragraph{Object Detection}
The pioneering work that introduced a transformer-based object detector was DETR \cite{DETR}, which demonstrated the ability to predict objects end-to-end without relying on NMS or other post-processing techniques. Subsequent research has built upon this foundation, aiming to enhance performance and address the limitations inherent in the original approach.

Deformable DETR \cite{Deformable_DETR} addressed the slow convergence and high computational cost of the original DETR by introducing deformable attention. DN-DETR \cite{DN_DETR} employed a denoising approach, introducing noisy targets as queries to improve prediction accuracy and speed up convergence. DAB-DETR \cite{DAB_DETR} further enhanced DETR by integrating dynamic anchor boxes into object queries, improving localization accuracy and guiding attention more effectively. Building on these advancements, DINO DETR \cite{DINO} refined key features of both DN-DETR and DAB-DETR and integrated RPN into DETR architecture. CO-DETR \cite{CO_DETR} incorporated classical object detectors, such as ATSS \cite{ATSS} and Faster R-CNN \cite{FRCNN}, into the DETR framework, leveraging the strengths of both paradigms.

\paragraph{Multi-Task Learning (MTL)}
Research in multimodal alignment, which integrates visual and textual modalities, has shown significant potential for improving performance in downstream tasks such as moment retrieval and highlight detection. Various methods have been proposed to tackle this challenge \cite{InternvideoV1, internvideo2}.

CLIP \cite{CLIP}, SimVLM \cite{Simvlm}, and OFA \cite{OFA}, which operate in the image domain, have demonstrated strong multimodal learning capabilities by leveraging contrastive and generative learning methods. However, their application to video-based tasks like moment retrieval remains suboptimal because they primarily focus on static images. It constrains their ability to capture the temporal dynamics inherent in video content.

Recent studies have introduced models that are specifically designed for video understanding tasks. For example, InternVideo \cite{InternvideoV1} utilized video masking and video-language contrastive learning as pretraining objectives, substantially improving video understanding. 
Further extending this approach, InternVideo2 \cite{internvideo2} employed a multi-stage training strategy that integrated self-supervised and weakly supervised techniques, including masked video token reconstruction and cross-modal contrastive learning. InternVideo2 demonstrated significant improvements across various downstream tasks, especially in the moment retrieval domain.
\begin{figure*}[ht!]
    \centering
    \includegraphics[width=0.95\textwidth]{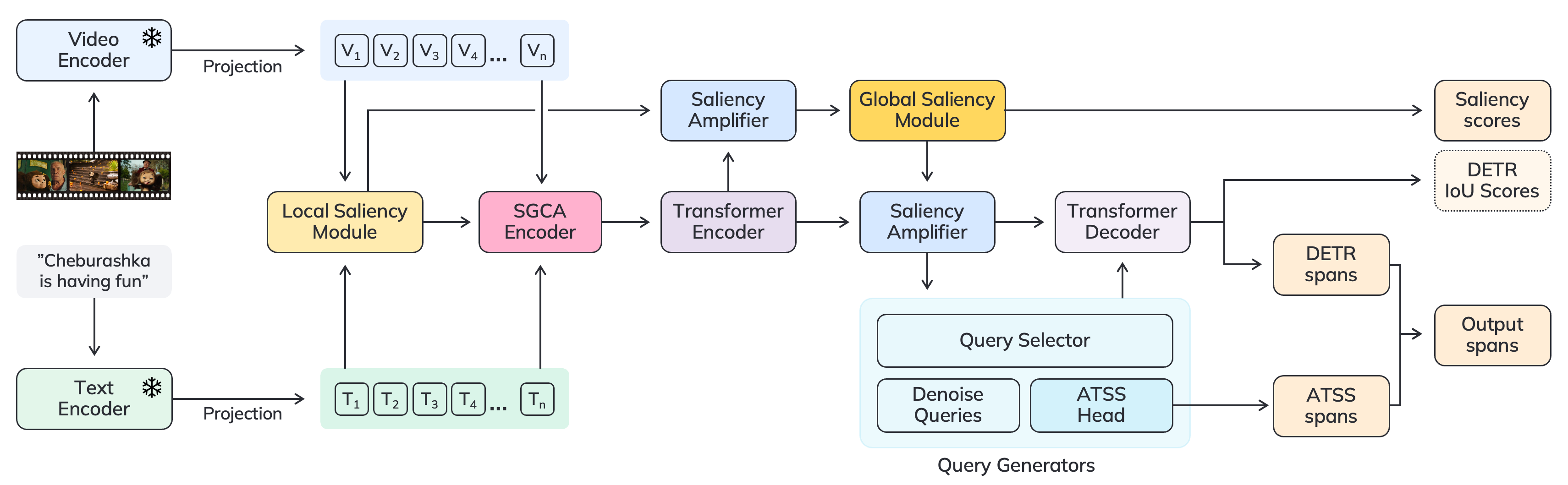}
    \caption{Saliency-Guided DETR architecture. Detailed explanations of notations are described in \cref{section:mr}.}
    \label{fig:Architecture}
\end{figure*}

\section{Method} 
\label{section:mr}
In this section, we describe each component of the proposed architecture, which effectively solves both tasks. An overview of the proposed Saliency-Guided DETR (SG-DETR) is shown in \cref{fig:Architecture}.

\subsection{Feature Extraction}
\label{subsection:extractor}
Previous studies \cite{QVHighlights, TR_DETR, QD_DETR, CG_DETR} have used a combination of the CLIP \cite{CLIP} and SlowFast \cite{SlowFast} models to extract features from the QVHighlights \cite{QVHighlights} dataset. They segmented the videos into 2-second intervals and extracted features from each segment using the CLIP and SlowFast models. These features were then concatenated.

This approach has several significant drawbacks. CLIP was initially trained to align images and texts. Thus, it does not fully account for the temporal dimension. This is why the SlowFast video extractor has to be added to the pipeline to compensate for the lack of spatial-temporal relationships in the CLIP features. However, the SlowFast and CLIP models were trained separately, so their embeddings are in different feature spaces.

Considering the abovementioned issues, we decided to use the InternVideoV2 \cite{internvideo2} model, which was trained for aligning video and text. It effectively handles the modeling of spatial-temporal relationships and does not require the inclusion of additional models like SlowFast in the pipeline, significantly simplifying it. 

Our feature extraction pipeline can be described as follows. We divide the video into 2-second clips. Then, we extract text and video features using the InternVideoV2-1B model. These features are then projected to a new dimension \(d\): \(F_v = [f_v^1, f_v^2, \ldots, f_v^L] \in \mathbb{R}^{L \times d}\) for video and \(F_t = [f_t^1, f_t^2, \ldots, f_t^N] \in \mathbb{R}^{N \times d}\) for text, where \(L\) and \(N\) denote the number of clips and textual tokens, respectively.


\subsection{Local Saliency Scores}
\label{subsection:local_saliency_scores}
Features obtained from the InternVideo2 encoders \cite{internvideo2} are already in a unified space, facilitating efficient and high-quality preliminary HD assessments. The scores obtained at this stage are termed as Local Saliency Scores $S_{\text{local}}$ because they do not take into account the global context of the entire video. To compute them, we first derive an overall vector of the text query using Pooling Encoder:
\begin{equation}\label{eq:attention_pooling}
    F_{\text{sent}} = \text{PoolingEncoder}(F_t),
\end{equation}
\noindent where PoolingEncoder is an encoder consisting of $N$ layers of AttentionPooling modules, and $F_{\text{sent}} \in \mathbb{R}^{1 \times d}$ represents the sentence token.

Next, we compute $S_{\text{local}}$ based on the similarity between the sentence token and the video embeddings. Formally, the local saliency scores are given by:
\begin{equation}\label{eq:cosine_similarity_eq}
    S_{\text{local}} = \alpha \cdot \text{L2-Norm}(F_{\text{sent}}) \cdot \text{L2-Norm}(F_v)^T + \beta,
\end{equation}
\noindent where $\alpha$ and $\beta$ are learnable scaling and shifting factors, respectively. The use of L2 normalization and affine transformation helps to stabilize the training process. To make $F_{\text{sent}}$ token even more informative we apply moment-sentence alignment as described in \cite{CG_DETR}.

\subsection{Saliency-Guided Cross Attention}
\label{subsection:sgca}

In standard cross-attention, softmax forces each video token to absorb some textual information, even when the correspondence is weak or irrelevant. As a result, video segments that don’t match the text query may still be influenced by it, degrading the model’s ability to distinguish relevant and irrelevant segments precisely. A possible solution is to replace softmax with sigmoid activations. However, using sigmoid breaks a text query's inter-token dependency, weakening the model's ability to rank relevant content \cite{CG_DETR}.

To address this issue, we introduce the Saliency-Guided Cross Attention (SGCA) mechanism, which effectively overcomes the limitations of standard cross-attention by considering the local relevance of each individual video clip to the text query. Formally, let the query tokens be derived from the video tokens as $Q = [p_Q(f_v^1), p_Q(f_v^2), \ldots, p_Q(f_v^L)]$, key and value tokens be derived from the text tokens: $K = [p_K(f_t^1), p_K(f_t^2), \ldots, p_K(f_t^L)]$ and $V = [p_V(f_t^1), p_V(f_t^2), \ldots, p_V(f_t^L)]$, where $p_Q(\cdot)$, $p_K(\cdot)$, and $p_V(\cdot)$ are learned projection functions that map the video and text tokens to a common hidden dimension $h$.

SGCA for the \( i \)-th video token \( f_v^i \) is computed as:

\begin{equation}\label{eq:sg_cross_attention}
    \text{SGCA}(f_v^i) = \sum_{j=1}^{L} W_{i,j} \odot V_j,
\end{equation}
where $W_{i,j}$ represents the attention weight for aligning the i-th video
token with the j-th text token and computed as

\begin{equation}\label{eq:weights}
    W_{i,j} = \text{Softmax}\left(\frac{Q_i \odot K_j}{\sqrt{h}}\right) \cdot \sigma(S_{\text{local}}^j)
\end{equation}
The proposed SGCA mechanism imposes only a minor computational burden, as detailed in Appendix A.

The described attention mechanism is then used as a part of the Cross-Attention Transformer to derive the video token representation enriched with textual information. After that, the standard Transformer Encoder is applied \cite{QD_DETR}.
\begin{equation}
    \begin{aligned}
        {F_v}^T &= \text{CrossModalEncoder}(F_v, F_t) \\
        \widehat{{F_v}^T} &= \text{TransformerEncoder}({F_v}^T)
    \end{aligned}
\end{equation}

\subsection{From Local to Global Saliency scores}
\label{subsection:gsm}
The local saliency scores can be further refined based on $\widehat{{F_v}^T}$. To do that, we first incorporate the information from the computed local scores into $\widehat{{F_v}^T}$ using the Saliency Amplifier (SA) module, which can be described as follows:

\begin{equation}\label{eq:local_saliency_ampl}
    \widehat{{{F_v}^T}_l} = \widehat{{F_v}^T} + \widehat{{F_v}^T} \cdot \sigma(S_{\text{local}}),
\end{equation}

These features are used to compute the offsets for the local saliency scores to get the global saliency scores:

\begin{equation}\label{eq:global_saliency_scores}
S_{\text{global}} = S_{\text{local}} + \text{Linear}(\widehat{{{F_v}^T}_l}).
\end{equation}

We reapply the SA module once the global saliency scores are obtained. The goal is to leverage the information obtained during the HD task to address the MR task.

\begin{equation}\label{eq:global_saliency_ampl}
\widehat{{{F_v}^T}_g} = \widehat{{F_v}^T} + \widehat{{F_v}^T} \cdot \sigma(S_{\text{global}}),
\end{equation}

We sent globally amplified features $\widehat{{{F_v}^T}_g}$ to regression heads.

\subsection{Hybrid Detector}
\label{subsection:hdetr}
In previous works, various detection heads, including CNN-based \cite{UniVTG} and DETR-like models \cite{QD_DETR, CG_DETR, QVHighlights}, have tackled the MR task. Co-DETR \cite{CO_DETR} has demonstrated that combining these approaches can be adequate for object detection tasks in the image domain. Inspired by this, we employ a hybrid detector with two detection heads. The ATSS \cite{ATSS} detector serves as the auxiliary CNN-based head, while the DINO-DETR \cite{DINO} detector is used as the primary detection head. In line with Co-DETR, we utilize positive anchors from the ATSS detector as queries for the DETR head during model training. Consequently, during training, the DETR head receives three sets of references: a few groups of noisy target spans, positive anchors from ATSS, and primary references generated by the query selector mechanism introduced in DINO-DETR. As part of the post-processing stage, we follow \cite{wbf} and apply a weighted fusion of ATSS and DETR boxes to produce the final predictions.

\subsection{Localization confidence}
\label{subsection:iou}
Prior research, including works such as \cite{IOUNET, retinaIOU, IOUdecoupled, AL_DETR}, has demonstrated that predicting IoU scores or a similar localization confidence metric significantly improves object detection performance. Building on these insights, we introduce an additional DETR head dedicated to predicting IoU scores between the predicted and ground truth spans. 

We incorporate predicted IoU scores into the matching process. Additionally, predicting IoU scores enables a better ranking of the predicted spans. Similar to the work \cite{retinaIOU}, we computed confidence of the span as a combination of classification and localization confidences: 

\begin{equation}\label{eq:conf_score} 
S_{det} = p_i^{\gamma} \hat{IoU_i}^{(1-\gamma)}
\end{equation}

where $p_i$ is the classification confidence, $\hat{IoU_i}$ is the predicted IoU, and ${\gamma}$ is a hyperparameter that controls the balance between $p_i$ and $\hat{IoU_i}$.

\section{Experiments and Results} 
\label{section: experiments}

\begin{table*}[ht]
\centering
\small
\begin{tabular}{lcccccc}
\hline
\multirow{2}{*}{Method} & \multirow{2}{*}{Features} & \multicolumn{3}{c}{MR mAP} & \multicolumn{2}{c}{HD $\ge$ Very Good} \\
\cmidrule(lr){3-5} \cmidrule(lr){6-7}
& & @0.5 & @0.75 & Avg. & mAP & HIT@1 \\
\hline
\multirow{2}{*}{Moment-DETR \cite{QVHighlights}} & CLIP+SlowFast & - & - & 32.20 & 35.65 & 55.55 \\
& InternVideo2-1b & 60.20 $\pm$ 0.55 & 34.43 $\pm$ 0.43 & 35.40 $\pm$ 0.41 & 40.31 $\pm$ 0.21 & 63.89 $\pm$ 0.62 \\
\hline
\multirow{2}{*}{UniVTG \cite{UniVTG}} & CLIP+SlowFast & - & - & 36.13 & 38.83 & 61.81 \\
& InternVideo2-1b & 63.51 $\pm$ 0.25 & 38.83 $\pm$ 0.26 & 37.78 $\pm$ 0.16 & 42.68 $\pm$ 0.09 & 69.34 $\pm$ 0.23 \\
\hline
\multirow{2}{*}{QD-DETR \cite{QD_DETR}} & CLIP+SlowFast & 62.23 & 41.82 & 41.22 & 39.13 & 63.03 \\
& InternVideo2-1b & 67.78 $\pm$ 0.29 & 46.40 $\pm$ 0.26 & 45.52 $\pm$ 0.15 & 41.82 $\pm$ 0.07 & 68.06 $\pm$ 0.24 \\
\hline
\multirow{2}{*}{CG-DETR \cite{CG_DETR}} & CLIP+SlowFast & 65.60 & 45.70 & 44.90 & 40.80 & 66.70 \\
& InternVideo2-1b & 69.86 $\pm$ 0.21 & 49.35 $\pm$ 0.28 & 48.69 $\pm$ 0.17 & 42.72 $\pm$ 0.07 & 69.87 $\pm$ 0.15 \\
\hline
\multirow{2}{*}{TR-DETR \cite{TR_DETR}} & CLIP+SlowFast & 66.27 & 46.42 & 45.09 & 40.55 & 64.77 \\
& InternVideo2-1b & 70.08 $\pm$ 0.15 & 49.20 $\pm$ 0.50 & 47.99 $\pm$ 0.42 & 43.43 $\pm$ 0.16 & 71.13 $\pm$ 0.25 \\
\hline
\rowcolor{gray!25} $\text{\ModelName}$ & InternVideo2-1b & \textbf{73.52 $\pm$ 0.05} & \textbf{57.91 $\pm$ 0.13} & \textbf{55.64 $\pm$ 0.20} & \textbf{43.91 $\pm$ 0.14} & \textbf{71.47 $\pm$ 0.73} \\
\hline
\end{tabular} 
\caption{Comparison of models performance on QVHighlights val split using different feature extractors. Results are reported as mean $\pm$ standard deviation, averaged over three runs.}
\label{tab:comparison_new_features}
\end{table*}

\subsection{Datasets}
We evaluate the effectiveness of the proposed method using three main benchmarks: QVHighlights \cite{QVHighlights}, Charades-STA \cite{Charades}, TACoS \cite{TACoS}.

QVHighlights is the primary benchmark annotated for both MR and HD tasks. The dataset is relatively small and consists of approximately 10k videos with human-written text queries covering various topics, from everyday activities and travel in lifestyle vlog videos to social and political activities in news videos. Charades-STA is a dataset used exclusively for moment retrieval task evaluation. It includes 16k sentence-moment pairs from 10k videos of indoor activities with an average duration of 30 seconds. TACoS consists of 127 cooking videos with an average duration of 287 seconds and 19k sentence-moment pairs. This dataset is recognized as particularly challenging because the moments occupy only a tiny portion of the considerably long videos.

\subsection{Feature Extraction}
In earlier studies \cite{UMT, QVHighlights, QD_DETR, EaTR}, various feature extractors were used on the mentioned benchmarks. For the QVHighlights benchmark, a combination of CLIP \cite{CLIP} and SlowFast \cite{SlowFast} models was employed. The Charades-STA benchmark used the VGG \cite{VGG} extractor for video content and GLOVE \cite{Glove} embeddings for text queries. This variety in methods requires maintaining multiple feature extraction pipelines and tuning hyperparameters for each, which slows research and increases the carbon footprint. Many of these methods have also become outdated and hardly remain representative. We propose a unified approach to address the problem. Specifically, we extracted visual and textual features for each benchmark using only the InterVidV2-1b \cite{internvideo2} model. We retrained some existing models using the InterVidV2-1b embeddings to assess the new features' impact and validate our method's competitiveness on the QVHighlights validation set. Although CLIP+SlowFast features allow additional comparisons with prior methods, they lack the necessary alignment for our SGCA module’s text–video similarity computation, leading to unstable training and reduced performance. The results are presented in \cref{tab:comparison_new_features}.

\subsection{Pretraining Framework}
\subsubsection{Motivation}
The QVHighlights dataset contains only 10k training samples. Previous studies \cite{QVHighlights, UniVTG} have shown that this dataset size is insufficient for training a model to effectively address a complex task like Moment Retrieval. As a result, even training on a noisy pretrain dataset can significantly improve the model's metrics. For instance, pretraining framework developed in \cite{UniVTG} boosted the model performance from 35.47 to 43.63 mAP@avg. Nevertheless, the metrics in the zero-shot mode remained relatively low at only 10.87 mAP@avg \cite{UniVTG}, leading us to believe that creating a dataset for pretraining remains an unsolved challenge.

\subsubsection{InterVid-MR Dataset}
Recent advancements in video-language modeling have been driven by the annotation of large scale, high-quality datasets like InternVid \cite{InternvideoV1}. Unlike VideoCC \cite{VideoCC} and other datasets previously used for pretraining \cite{Ego4D, QVHighlights}, InternVid was annotated using LLMs and does not rely on pseudo-labeling, resulting in significantly less annotation noise. To construct our dataset, we used the subset InternVid-10M-FLT \cite{InternvideoV1} derived from the original InternVid.

Unfortunately, InternVid-10M-FLT can not be directly used to train MR models because each caption is linked to only one video segment in the annotations, whereas in practice, multiple video segments may closely match a given query. We have made several steps to address this issue and developed a new dataset named $\textbf{InterVid-MR}$, which is suitable for MR and HD tasks.

First, we additionally filtered the captions of the InternVid-10M-FLT dataset to remove meaningless ones using ChatGPT \footnote{https://chat.openai.com/chat}. Next, to align the data with the QVHighlights dataset, we selected video segments no longer than 150 seconds for each caption from the InternVid-10M-FLT dataset. These segments were created by trimming the original videos around the time intervals associated with the captions.

Then, we split each video into 2-second clips and extracted features using the IntervidV2 video encoder. Caption features were extracted using the IntervidV2 text encoder. We computed the cosine similarity between the embeddings of each video clip and the corresponding caption embedding using the formula:
\begin{equation}\label{eq:cosine_similarity}
    S = \text{L2\_Norm}(F_v) \cdot \text{L2\_Norm}(F_t)^T,
\end{equation}

\noindent where $F_v$ represents the embedding of the video clip, and $F_t$ represents extracted cls embedding of the caption.

We used the similarity scores of video clips within the original positive interval, defined by the dataset annotation, to determine the minimum similarity score $S_{\text{min}}$ required for a clip to be included in a new positive interval. This minimum similarity score $S_{\text{min}}$ is calculated as:
\begin{equation} \label{eq:min_score}
    S_{\text{min}} = \operatorname{mean}(S_{\text{pos}}) - 3 \times \operatorname{std}(S_{\text{pos}})
\end{equation}

\noindent where $S_{\text{pos}}$ represents the similarity scores of all clips within the original positive interval. Based on the $S_{\text{min}}$ threshold, we constructed new positive intervals consisting of clips whose similarity scores are higher than $S_{\text{min}}$.

After annotating the MR task, we proceeded to annotate the HD task. We used the similarity scores calculated in the previous step to assign each clip within the identified positive intervals a grade ranging from 1 to 4 following the QVHighlights annotation scheme \cite{QVHighlights}. The grades are calculated using the following formula:
\begin{equation}\label{eq:score_function}
\text{Grade}(S) =
\begin{cases}
    4, & \text{if } S \in (\mu + 1.5 \times \sigma, S_{\text{max}}] \\
    3, & \text{if } S \in (\mu, \mu + 1.5 \times \sigma] \\
    2, & \text{if } S \in (\mu - 1.5 \times \sigma, \mu] \\
    1, & \text{if } S \in [S_{\text{min}}, \mu - 1.5 \times \sigma]
\end{cases}
\end{equation}
\noindent where $S$ is the score from the positive interval to be graded, $\mu$ is the mean score of the positive clips, $\sigma$ is the standard deviation of the positive clips' scores, and $S_{\text{max}}$ is the maximum score of the positive clips. The constants in \cref{eq:min_score} and \cref{eq:score_function} were determined empirically. Visualization of the annotation creation process for pretraining is presented in Appendix C.

As a result, we obtain a dataset that can be used to solve both MR and HD tasks simultaneously.

\subsubsection{Impact of the Pretraining}
Using the proposed data annotation framework, we collected a dataset consisting of 150k samples. To ensure the effectiveness of the proposed data annotation strategy for solving MR and HD tasks, we measured the zero-shot performance of a model trained on different amounts of pre-train data on the QVHighlights dataset. As shown in \cref{fig:pretrain_impact}, even 20k samples in the training dataset are sufficient to achieve an MR mAP@avg above 45. Using all the data available, we surpassed the 50 MR mAP@avg threshold, which significantly exceeds the results of the pretraining strategy proposed in \cite{UniVTG}.

\begin{figure}
    \centering
    \begin{tikzpicture}
        \definecolor{pastelblue}{rgb}{0.2, 0.2, 0.2}
        \definecolor{pastelred}{rgb}{1.0, 0.71, 0.76}
        \definecolor{pastelgreen}{rgb}{0.6, 0.98, 0.6}

        \begin{axis}[
            xlabel={Number of Pretrain Samples in thousands},
            ylabel={mAP@avg (\%)},
            xmin=30, xmax=150,
            ymin=45, ymax=63,
            xtick={30, 60, 90, 120, 150},
            ytick={45, 50, 55, 60},
            legend pos=south east,
            grid=major,
            thick,
            width=0.470\textwidth,
            height=0.285\textwidth,
            axis line style={line width=0.25pt},
            label style={font=\footnotesize},
            tick label style={font=\footnotesize},
            legend style={font=\footnotesize, at={(0.97, 0)}, draw=none},
            legend columns=3,
            legend cell align={left},
            ymajorgrids=true,
            xmajorgrids=true
        ]

        \addplot[
            color=pastelblue,
            mark=none,
            dashed,
            thick
        ] coordinates {
            (30,55.64) (60,55.64) (90,55.64) (120,55.64) (150,55.64)
        };
        \addlegendentry{QVHighlights}

        \addplot[
            color=pastelgreen,
            mark=square*,
            thick
        ] coordinates {
            (30,47.95) (60,49.12) (90,49.87) (120,50.30) (150,50.31)
        };
        \addlegendentry{Pretrain}

        \addplot[
            color=pastelred,
            mark=*,
            thick
        ] coordinates {
            (30,57.8) (60,59.3) (90,60.0) (120,60.5) (150,59.9)
        };
        \addlegendentry{Finetune}

        \end{axis}
    \end{tikzpicture}
    \caption{The impact of pre-train dataset size on MR mAP@avg metric on the QVHighlights validation set.}
    \label{fig:pretrain_impact}
\end{figure}

\begin{table*}[htbp]
\centering
\small
\setlength{\tabcolsep}{3pt}
\renewcommand{\arraystretch}{1.1}
\begin{tabular}{cccccccccccc}
\toprule
 & & & & & \multicolumn{5}{c}{MR} & \multicolumn{2}{c}{HD} \\
 \cmidrule(lr){6-10} \cmidrule(lr){11-12}
 & SGCA & IoU Scoring & GSM & Hybrid DETR & \multicolumn{2}{c}{R1} & \multicolumn{3}{c}{mAP} & \multicolumn{2}{c}{$\geq$ Very Good} \\
 \cmidrule(lr){6-7} \cmidrule(lr){8-10} \cmidrule(lr){11-12}
 & & & & & @0.5 & @0.7 & @0.5 & @0.75 & Avg. & mAP & HIT@1 \\
 \midrule
(a) &  &  &  &  & 70.1 $\pm$ 0.3 & 53.7 $\pm$ 0.5 & 70.0 $\pm$ 0.1 & 49.3 $\pm$ 0.2 & 48.3 $\pm$ 0.4 & 41.9 $\pm$ 0.1 & 68.6 $\pm$ 0.8 \\
(b) & \checkmark & &  &  & 71.1 $\pm$ 0.5 & 56.5 $\pm$ 0.8  & 71.0 $\pm$ 0.4 & 52.2 $\pm$ 0.7 & 50.1 $\pm$ 0.6 & 42.2 $\pm$ 0.1 & 70.1 $\pm$ 0.6 \\
(c) & \checkmark & \checkmark &  &  & 70.8 $\pm$ 0.2 & 56.5 $\pm$ 0.5 & 71.3 $\pm$ 0.3 & 52.5 $\pm$ 0.3 & 51.1 $\pm$ 0.2 & 42.3 $\pm$ 0.1 & 69.9 $\pm$ 0.5 \\
(d) & \checkmark & \checkmark & \checkmark &  & 72.1 $\pm$ 0.9 & 57.6 $\pm$ 0.4 & 72.6 $\pm$ 0.3 & 53.6 $\pm$ 0.5 & 52.2 $\pm$ 0.3 & 43.8 $\pm$ 0.2 & 71.1 $\pm$ 0.4 \\
(e) & \checkmark & \checkmark & \checkmark & \checkmark & \textbf{72.8 $\pm$ 0.5}  & \textbf{59.5 $\pm$ 0.2} & \textbf{73.5 $\pm$ 0.1} & \textbf{57.9 $\pm$ 0.1} & \textbf{55.6 $\pm$ 0.2} & \textbf{43.9 $\pm$ 0.1} & \textbf{71.4 $\pm$ 0.7} \\
\bottomrule
\end{tabular}
\caption{Ablation study on QVHighlights val split. SGCA and GSM stand for saliency-guided cross-attention transformer encoder and global saliency module, respectively. Results are reported as mean $\pm$ standard deviation, averaged over three runs.}
\label{tab:abl_stad}
\end{table*}

\subsection{Training Details} 
We train $\text{\ModelName}$ using two attention-pooling layers in the Local Saliency Head and two cross-attention transformer layers in the SGCA Encoder. The model architecture includes three layers of encoders and three layers of decoders. The hidden dimension of the transformer layers is set to 256. We employ the AdamW optimizer \cite{adamw} with a weight decay of 1e-1. A dropout rate of 0.1 is applied throughout the model, with a higher dropout rate of 0.5 in the projection layers. Our query selector mechanism generates 25 queries per instance. Unless otherwise specified, the following loss balancing parameters are used:
$\lambda_{\text{L1}} = 10$, 
$\lambda_{\text{gIoU}} = 1$, 
$\lambda_{\text{CE}} = 5$,
$\lambda_{\text{Centerness}} = 1$,
$\lambda_{\text{detr}} = 1$, 
$\lambda_{\text{atss}} = 1$, 
$\lambda_{\text{hl}} = 1$,
$\lambda_{\text{aux}} = 1$.
A detailed description of each loss component is provided in Appendix D.

\noindent Dataset-Specific Settings:
\begin{itemize}
    \item \textbf{QVHighlight:} The model is trained on 1xV100 GPU for 150 epochs with a batch size of 128 and a learning rate (lr) of 5e-4. We employ the WarmupMultiStepLR scheduler, utilizing 45 warmup steps, with two milestones set at 100 and 125 epochs. The lr is decayed by a factor of 0.5 at each milestone.    
    \item \textbf{Charades-STA:} The model is trained on 1xV100 GPU for 100 epochs with a batch size of 64 and a lr of 5e-4. Video features are extracted from 1-second clips, as it was done in \cite{BAM_DETR}. During training, binary highlight scores were generated based on whether a clip overlapped more than halfway with a relevant moment retrieval interval.
    \item \textbf{TACoS:} The model is trained on 1xV100 GPU for 120 epochs with a batch size of 32 and a lr of 5e-4. Video features are extracted from 2-second, as it was done in \cite{BAM_DETR}. Binary highlight scores were generated the same way as in Charades-STA.
\end{itemize}

\begin{table}[ht]
\centering
\small
\setlength{\tabcolsep}{1pt}
\renewcommand{\arraystretch}{0.95} 
\small
\begin{tabular}{lcccccccc}
\toprule
& \multicolumn{5}{c}{MR} & \multicolumn{2}{c}{HD} \\
\cmidrule(lr){2-6} \cmidrule(lr){7-8}
Method & \multicolumn{2}{c}{R1} & \multicolumn{3}{c}{mAP} & \multicolumn{2}{c}{$\ge$ Very Good} \\
\cmidrule(lr){2-3} \cmidrule(lr){4-6} \cmidrule(lr){7-8}
& @0.5 & @0.7 & @0.5 & @0.75 & Avg. & mAP & HIT@1 \\
\midrule
MDETR \cite{QVHighlights} & 52.9 & 33.0 & 54.8 & 29.4 & 30.7 & 35.7 & 55.6 \\
UMT † \cite{UMT} & 56.2 & 41.2 & 53.4 & 37.0 & 36.1 & 38.2 & 60.0 \\
UniVTG \cite{UniVTG} & 58.9 & 40.9 & 57.6 & 35.6 & 35.5 & 38.2 & 61.0 \\
QD-DETR \cite{QD_DETR} & 62.4 & 45.0 & 62.6 & 39.9 & 39.9 & 38.6 & 62.4 \\
CG-DETR \cite{CG_DETR} & 65.4 & 48.4 & 64.5 & 42.8 & 42.9 & 40.3 & 66.2 \\
BAM-DETR \cite{BAM_DETR} & 62.7 & 48.6 & 64.6 & 46.3 & 45.4 & - & - \\
TR-DETR \cite{TR_DETR} & 64.7 & 49.0 & 64.0 & 43.7 & 42.6 & 39.9 & 63.4 \\
Mr. BLIP \cite{MrBLIP} & \textbf{74.7} & \textbf{60.5} & 68.1 & 53.4 & 51.4 & - & - \\
\rowcolor{gray!25} $\text{\ModelName}$ & 72.2 & 56.6  & 73.2 & 55.8 & 54.1 & 43.8 & 69.1 \\
\rowcolor{gray!25} $\text{\ModelName}$ ZS & 63.9 & 49.6  & 67.5 & 49.0 & 48.3 & 43.0 & 68.0 \\
\rowcolor{gray!25} $\text{\ModelName}$ w/ PT & 74.2 & 60.4  & \textbf{76.2} & \textbf{60.8} & \textbf{58.8} & \textbf{44.7} & \textbf{71.0} \\
\hline
\end{tabular}
\caption{Experimental results on the QVHighlights \textit{test} set. w/ PT means fine-tuning after pre-training; ZS means zero-shot inference after pre-training. † denotes using audio modality. $\text{\ModelName}$ uses InternVideo2-1b, MR. BLIP uses BLIP-2, and other models use CLIP+SlowFast for FE.}
\label{tab:comparison_existing_features}
\end{table}

\begin{table}[ht]
\centering
\small
\setlength{\tabcolsep}{2pt}
\renewcommand{\arraystretch}{0.95}
\begin{tabular}{lccccccc}
\toprule
& \multicolumn{3}{c}{Charades-STA} & \multicolumn{3}{c}{TACoS} \\
\cmidrule(lr){2-4} \cmidrule(lr){5-7}
Method & R@0.5 & R@0.7 & mIoU & R@0.5 & R@0.7 & mIoU \\
\midrule
MDETR \cite{QVHighlights}     & 53.6 & 31.4 & -    & 24.7 & 12.0 & 25.5 \\
UMT † \cite{UMT}               & 48.3 & 29.3 & -    & -    & -    & -    \\
UniVTG \cite{UniVTG}          & 58.0 & 35.7 & 50.1 & 35.0 & 17.4 & 33.6 \\
QD-DETR \cite{QD_DETR}        & 57.3 & 32.6 & -    & 36.8 & 21.1 & 35.8 \\
CG-DETR \cite{CG_DETR}        & 58.4 & 36.3 & 50.1 & 39.6 & 22.2 & 36.5 \\
BAM-DETR \cite{BAM_DETR}      & 60.0 & 39.4 & 52.3 & 41.5 & 26.8 & 39.3 \\
TR-DETR \cite{TR_DETR}        & 57.6 & 33.5 & -    & - & - & - \\
MR. BLIP \cite{MrBLIP}        & 69.3 & 49.3 & 58.6 & - & - & - \\
\rowcolor{gray!25} $\text{\ModelName}$       & 70.2 & 49.5 & 59.1 & 44.7 & 29.9 & 40.9 \\
\rowcolor{gray!25} $\text{\ModelName}$ w/ PT & \textbf{71.1} & \textbf{52.8} & \textbf{60.7} &  \textbf{46.4} &  \textbf{33.9} &  \textbf{42.4} \\
\hline
\end{tabular}
\caption{Experimental results on the Charades-STA and TACoS \textit{test} sets. w/ PT means fine-tuning after pre-training. † denotes using audio modality. $\text{\ModelName}$ uses InternVideo2-1b, MR. BLIP uses BLIP-2, and other models use CLIP+SlowFast for FE.}
\label{tab:charades}
\end{table}

\subsection{Comparison with the state-of-the-art}

We compare $\text{\ModelName}$ against existing state-of-the-art methods on both joint and single-task benchmarks. This quantitative analysis is complemented by a qualitative visualization of our model's predictions in Appendix B.

\paragraph{Joint Moment Retrieval and Highlight Detection}
To evaluate the effectiveness of $\text{\ModelName}$ in jointly addressing MR and HD tasks, we conducted experiments on the QVHighlights test split, as shown in \cref{tab:comparison_existing_features}. Remarkably, even without pretraining, $\text{\ModelName}$ significantly outperforms all existing methods that do not leverage LLMs, and it delivers performance comparable to the MrBLIP model despite the latter having considerably more parameters. After pretraining on our constructed InterVid-MR dataset, $\text{\ModelName}$ achieves competitive zero-shot performance, with 48.3 mAP and 68.0 HIT@1 results. When fine-tuned on the QVHighlights dataset, the pre-trained $\text{\ModelName}$ surpasses all existing methods, including those based on LLMs, with 58.8 mAP and 71.0 HIT@1 results.


We also evaluate on two small-scale highlight detection benchmarks, TVSum \cite{TVSUM} and YouTube Highlights \cite{Youtube}. As detailed in \cref{tab:highlight_detection_results_tvsum} and \cref{tab:youtube}, our model achieves state-of-the-art results on YouTube Highlights. In contrast, its performance on the data-scarce TVSum is constrained; we attribute this primarily to the dataset's limited size being insufficient for stable model training. Consequently, fine-tuning with supplemental data does not yield further performance gains on this benchmark.

\begin{table}[h]
\centering
\footnotesize
\setlength{\tabcolsep}{4pt}
\begin{tabular}{lccccccc}
\hline
Method & Dog & Gym. & Par. & Ska. & Ski. & Sur. & Avg. \\
\hline
UMT†  \cite{UMT}          & 65.9 & 75.2 & 81.6 & 71.8 & 72.3 & 82.7 & 74.9 \\
UniVTG \cite{UniVTG} & 71.8 & 76.5 & 73.9 & 73.3 & 73.2 & 82.2 & 75.2 \\
QD-DETR \cite{QD_DETR} & 72.2 & 77.4 & 71.0 & 72.7 & 72.8 & 80.6 & 74.4 \\
CG-DETR \cite{CG_DETR} & \textbf{76.3} & 76.1 & 76.0 & 75.1 & \textbf{81.9} & 75.9 & 75.9 \\
\hline
\textbf{\ModelName{}}       & 75.1 & \textbf{79.7} & 76.9 & 71.0 & 74.5 & \textbf{83.2} & 76.7 \\
\textbf{\ModelName{}} ZS       & 65.0 & 69.1 & 76.6 & 57.5 & 76.1 & 70.8 & 69.2 \\
\textbf{\ModelName{}} w/ PT & 66.7 & 78.3 & \textbf{86.5} & \textbf{78.6} & 75.9 & 82.2 & \textbf{78.0} \\
\hline
\end{tabular}
\caption{Experimental results on the Youtube-HL. The mAP is used as a metric. † denotes using audio modality. w/ PT means fine-tuning after pre-training; ZS means zero-shot inference after pre-training, \textbf{bold} letters indicate the best results.}
\label{tab:youtube}
\end{table}

\begin{table}[h]
\centering
\footnotesize
\setlength{\tabcolsep}{2pt}
\renewcommand{\arraystretch}{1.1}
\resizebox{0.49\textwidth}{!}{
\begin{tabular}{lccccccccccccc}
\hline
Method & VT & VU & GA & MS & PK & PR & FM & BK & BT & DS & Avg. \\
\hline
UMT† \cite{UMT}        & 87.5 & 81.5 & 88.2 & 78.8 & 81.5 & 87.0 & 76.0 & 86.9 & 84.4 & 79.6 & 83.1 \\
UniVTG \cite{UniVTG}   & 83.9 & 85.1 & 89.0 & 80.1 & 84.6 & 81.4 & 70.9 & 91.7 & 73.5 & 69.3 & 81.0 \\
QD-DETR \cite{QD_DETR} & 88.2 & 87.4 & 85.6 & 85.0 & 85.8 & 86.9 & 76.4 & 91.3 & 89.2 & 73.7 & 85.0 \\
CG-DETR \cite{CG_DETR} & 86.9 & 88.8 & \textbf{94.8} & \textbf{87.7} & 86.7 & \textbf{89.6} & 74.8 & \textbf{93.3} & 89.2 & 75.9 & 86.8 \\
TR-DETR \cite{TR_DETR} & 89.3 & \textbf{93.0} & 94.3 & 85.1 & 88.0 & 88.6 & \textbf{80.4} & 91.3 & 89.5 & \textbf{81.6} & \textbf{88.1} \\
\hline
\textbf{\ModelName{}} & \textbf{89.9} & 90.5 & 88.0 & 87.4 & 90.0 & 84.5 & 77.9 & 93.2 & \textbf{89.9} & 80.3 & 87.1 \\
\textbf{\ModelName{}} ZS & 44.3 & 70.9 & 85.5 & 65.7 & 61.1 & 49.0 & 44.3 & 67.3 & 83.2 & 48.8 & 62.0 \\
\textbf{\ModelName{}} w/ PT & 86.6 & 74.2 & 85.9 & 82.9 & \textbf{90.6} & 74.5 & 69.0 & 92.6 & 87.6 & 67.7 & 81.2 \\
\hline
\end{tabular}
}
\caption{Experimental results on the TVSum. The Top-5 mAP is used as a metric. † denotes using audio modality. w/ PT means fine-tuning after pre-training; ZS means zero-shot inference after pre-training, \textbf{bold} letters indicate the best results.}
\label{tab:highlight_detection_results_tvsum}
\end{table}

\paragraph{Moment Retrieval} \vspace{-5pt}
We use the Charades-STA and TACoS datasets as benchmarks to evaluate the performance of $\text{\ModelName}$ in the moment retrieval task. The results are presented in \Cref{tab:charades}. $\text{\ModelName}$ achieves state-of-the-art performance on these benchmarks, both with and without using additional data from our pretraining framework. The results highlight the model's robustness and its ability to generalize effectively across varying task conditions.

\subsection{Ablation study}
To evaluate the contribution of each component in the proposed approach, we conducted an ablation study, as shown in \cref{tab:abl_stad}. The baseline model (a) utilizes a standard cross-attention transformer for text-to-vision interaction \cite{QD_DETR}, a DETR-based regression head \cite{QVHighlights, UniVTG}, and the local saliency head described in \cref{subsection:local_saliency_scores}. We selected this architecture as our baseline due to its simplicity and reliance on well-established components. In experiment (b), we assess the effect of incorporating the SGCA encoder, detailed in \cref{subsection:sgca}. It increases model convergence and pushes performance forward, especially the moment retrieval part. Experiment (c) introduces the IoU scoring mechanism, described in \cref{subsection:iou}, resulting in growth of moment retrieval performance. Next, in experiment (d), we investigate the influence of the GSM module, discussed in \cref{subsection:gsm}, on performance. It improves not only MR performance but especially the quality of HD. Finally, in experiment (e), we replace the standard DETR head with a hybrid ATSS-DETR head, outlined in \cref{subsection:hdetr}, improving MR performance.

\section{Conclusions}
\label{section:conclusions}
We introduced $\text{\ModelName}$, a unified framework for Moment Retrieval and Highlight Detection, and presented an approach for generating pretraining data, enabling us to create the InterVid-MR dataset. Evaluations on benchmarks such as QVHighlights, Charades-STA, and TACoS show that $\text{\ModelName}$ sets state-of-the-art results across all tasks. Ablation studies highlight the effectiveness of key components like the Saliency-Guided Cross-Attention encoder and hybrid ATSS-DETR head, offering a scalable and robust solution for video-language modeling.

\section{Limitations}
\label{section:limitations}
We construct the InterVid-MR dataset using annotations generated by LLMs. Misalignments between captions produced by LLMs and their corresponding video segments may introduce errors in video-text alignment, negatively impacting both model training and performance on fine-grained queries. Additionally, our hyperparameters were selected empirically, and further investigation could potentially yield improved results. Our method also excludes the audio modality, which is a decision motivated not only by modeling considerations but also by the characteristics of the datasets, as QVHighlights and InterVid-MR primarily focus on visual content and lack aligned audio annotations.


{
    \small
    \bibliographystyle{ieeenat_fullname}
    \bibliography{main}
}

\clearpage
\appendix

\twocolumn[
\centering
{\LARGE\bfseries Supplementary Material}\par
\vspace{1.5em}
]


\begin{table*}[!t]
\centering
\small
\setlength{\tabcolsep}{3pt}
\renewcommand{\arraystretch}{1.1}
\begin{tabular}{ccccccccc}
\toprule
 & & \multicolumn{5}{c}{MR} & \multicolumn{2}{c}{HD} \\
 \cmidrule(lr){3-7} \cmidrule(lr){8-9}
 & Attention Type & \multicolumn{2}{c}{R1} & \multicolumn{3}{c}{mAP} & \multicolumn{2}{c}{$\geq$ Very Good} \\
 \cmidrule(lr){3-4} \cmidrule(lr){5-7} \cmidrule(lr){8-9}
 & & @0.5 & @0.7 & @0.5 & @0.75 & Avg. & mAP & HIT@1 \\
 \midrule
(a) & CA & 70.1 $\pm$ 0.3 & 53.7 $\pm$ 0.5 & 70.0 $\pm$ 0.1 & 49.3 $\pm$ 0.2 & 48.3 $\pm$ 0.4 & 41.9 $\pm$ 0.1 & 68.6 $\pm$ 0.8 \\
(b) & ACA & 67.7 $\pm$ 0.6 & 51.5 $\pm$ 0.7  & 68.22 $\pm$ 0.2 & 47.7 $\pm$ 0.8 & 47.1 $\pm$ 0.7 & 42.0 $\pm$ 0.1 & 69.41 $\pm$ 0.5 \\
(c) & SGCA & 71.1 $\pm$ 0.5 & 56.5 $\pm$ 0.8  & 71.0 $\pm$ 0.4 & 52.2 $\pm$ 0.7 & 50.1 $\pm$ 0.6 & 42.2 $\pm$ 0.1 & 70.1 $\pm$ 0.6 \\
\bottomrule
\end{tabular}
\caption{Ablation study of attention types on QVHighlights val split. CA, ACA, and SGCA stand for Cross Attention, Adaptive Cross-Attention, and Saliency-Guided Cross-Attention. Results are reported as mean $\pm$ standard deviation, averaged over three runs.}
\label{tab:comparison_attn}
\end{table*}

\section{SG-Attention: Computational Efficiency}
\label{app:sgca_efficiency}

SG-Attention only minimally increases computational cost compared to standard dot-product attention. 
In the conventional setup, the main cost arises from two matrix multiplications: 
$QK^\top$ and $\mathrm{AttnWeights} \times V$. 
Each multiplication requires $2L^2 E$ multiply-add operations, leading to a total of $4L^2 E$. 
By contrast, SG-Attention performs one additional elementwise multiplication of size $L^2$, which is negligible when $E$ is large. For instance, with $L=76$ and $E=256$, standard attention incurs approximately $5.915 \times 10^6$ operations, 
whereas SG-Attention adds only about $5.776 \times 10^3$ operations (less than $1\%$ overhead). 
Consequently, this saliency mechanism imposes only a minor computational burden and does not impede real-time performance.

\section{Comparative Study of Attention Mechanisms}
\label{app:sgca_comp}
We evaluate and compare three distinct attention mechanisms for fusing textual information into video feature representations: standard Cross-Attention, the Adaptive Cross-Attention proposed by CGDETR \cite{CG_DETR}, and our novel Saliency-Guided Attention. The results for the baseline architecture, trained on the QVHighlights dataset with each of the three attention variants, are summarized in \cref{tab:comparison_attn}.
\section{Training Objectives}
\label{app:losses}
Our model is trained using losses categorized into three main groups.

\paragraph{Highlight Detection Task} \vspace{-6pt} We employ margin ranking, rank contrastive, and cross-entropy (CE) losses on both local and global saliency scores, as defined in \cite{CG_DETR}. Additionally, we apply the CE loss to negative text-video pairs following \cite{QD_DETR} to suppress negative clip saliency. The total loss for the task is represented as:
\vspace{-1pt}
\begin{equation}\label{eq:total_sal_loss} 
\mathcal{L}_{hl} = \mathcal{L}_{marg} + \mathcal{L}_{rctl} + \mathcal{L}_{bce\_pos}  + \mathcal{L}_{bce\_neg} 
\end{equation}

\paragraph{Moment Retrieval Task} \vspace{-6pt} For the MR task, we use CE loss, generalized IoU (GIoU) loss \cite{GIoU}, and smooth L1 loss to train both DETR and ATSS detection heads. For the ATSS head \cite{ATSS}, we also incorporate Centerness Loss \cite{FCOS} to enhance localization precision. In the DETR head, CE loss is applied to IoU head predictions. Losses for auxiliary DETR queries are computed similarly to primary queries. The objectives for the task are:

\begingroup
\vspace{-1em} 
\begin{equation}\label{eq:mr_loss_detr}
\begin{split}
\mathcal{L}_{\text{detr}} &= \lambda_{\text{L1}} \mathcal{L}_{\text{L1}}(m, \bar{m}) 
+ \lambda_{\text{gIoU}} \mathcal{L}_{\text{gIoU}}(m, \bar{m}) \\
&\quad + \lambda_{\text{CE}} \mathcal{L}_{\text{CE}}(y, \bar{y}) + \lambda_{\text{IoU}} \mathcal{L}_{\text{CE}}(IoU, \bar{IoU})
\end{split}
\end{equation}

\vspace{-1em} 
\begin{equation}\label{eq:mr_loss_atss}
\begin{split}
\mathcal{L}_{\text{atss}} &= \lambda_{\text{L1}} \mathcal{L}_{\text{L1}}(m, \bar{m}) 
+ \lambda_{\text{gIoU}} \mathcal{L}_{\text{gIoU}}(m, \bar{m}) \\
&\quad + \lambda_{\text{CE}} \mathcal{L}_{\text{CE}}(y, \bar{y}) 
+ \lambda_{\text{Centrness}} \mathcal{L}_{\text{CE}}(c, \bar{c})
\end{split}
\end{equation}
\endgroup

\noindent In this context, the ground truth values are represented as \( m = (m_c, m_\sigma) \), \( y \), \( c \), and \( IoU \), where \( m_c \) and \( m_\sigma \) denote the center and duration of the ground-truth moment, \( y \) represents the binary classification label, \( c \) is the target centerness score, and \( IoU \) refers to the target IoU score. Similarly, the predicted values are denoted as \( \bar{m} = (\bar{m}_c, \bar{m}_\sigma) \), \( \bar{y} \), \( \bar{c} \), and \( \hat{IoU_i} \), corresponding to the predicted moment, binary classification label, centerness score, and predicted IoU score, respectively.

\paragraph{Auxiliary Losses} \vspace{-5pt} The third category includes auxiliary losses to enhance the model's overall performance. First, alignment loss ensures consistency between moment and sentence token. Additionally, CE loss is applied to differentiate moment tokens from non-moment tokens within each video instance. Detailed descriptions of these losses can be found in \cite{CG_DETR}. The overall objective of the group is:
\begin{equation}\label{eq:aux_loss}
\mathcal{L}_{aux} = \mathcal{L}_{bce} + \mathcal{L}_{align}
\end{equation}

\paragraph{Overall Objective} \vspace{-5pt}
The final objective function is the sum of all the aforementioned losses:
\begin{equation}\label{eq:total_loss}
\mathcal{L}_{obj} = \lambda_{\text{atss}}\mathcal{L}_{\text{atss}} + \lambda_{\text{detr}}\mathcal{L}_{\text{detr}} + \lambda_{\text{hl}}\mathcal{L}_{hl} + \lambda_{\text{aux}}\mathcal{L}_{aux} 
\end{equation}

\section{Qualitative Results and Analysis}
\label{app:qualitative_results}

\begin{figure*}[!b]
    \centering
    \includegraphics[width=0.9\textwidth]{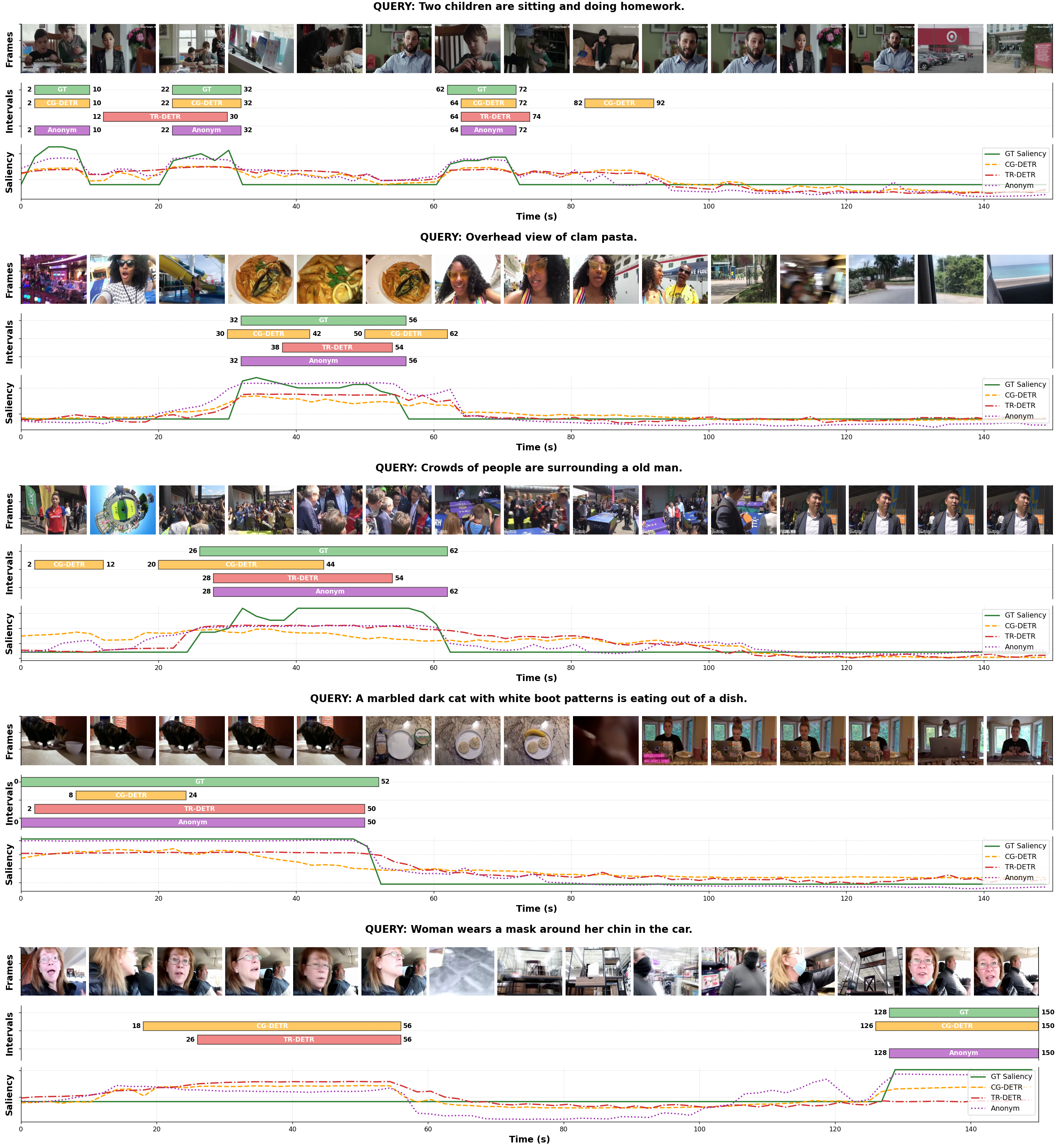}
    \caption{Qualitative results}
    \label{fig:QualityVis}
\end{figure*}

In \cref{fig:QualityVis}, we present visualizations of predictions made by the models on the QVHighlights validation dataset. Compared to existing methods such as CG-DETR \cite{CG_DETR} and TR-DETR \cite{TR_DETR}, $\text{\ModelName}$ achieves more precise and coherent highlight detection results, as evidenced by improvements in both retrieval accuracy and highlight score distributions.

\section{Visualization of the Pretrain Annotation Framework}
\label{app:visualization}

\begin{figure*}[ht]
    \centering
    \includegraphics[width=0.95\textwidth]{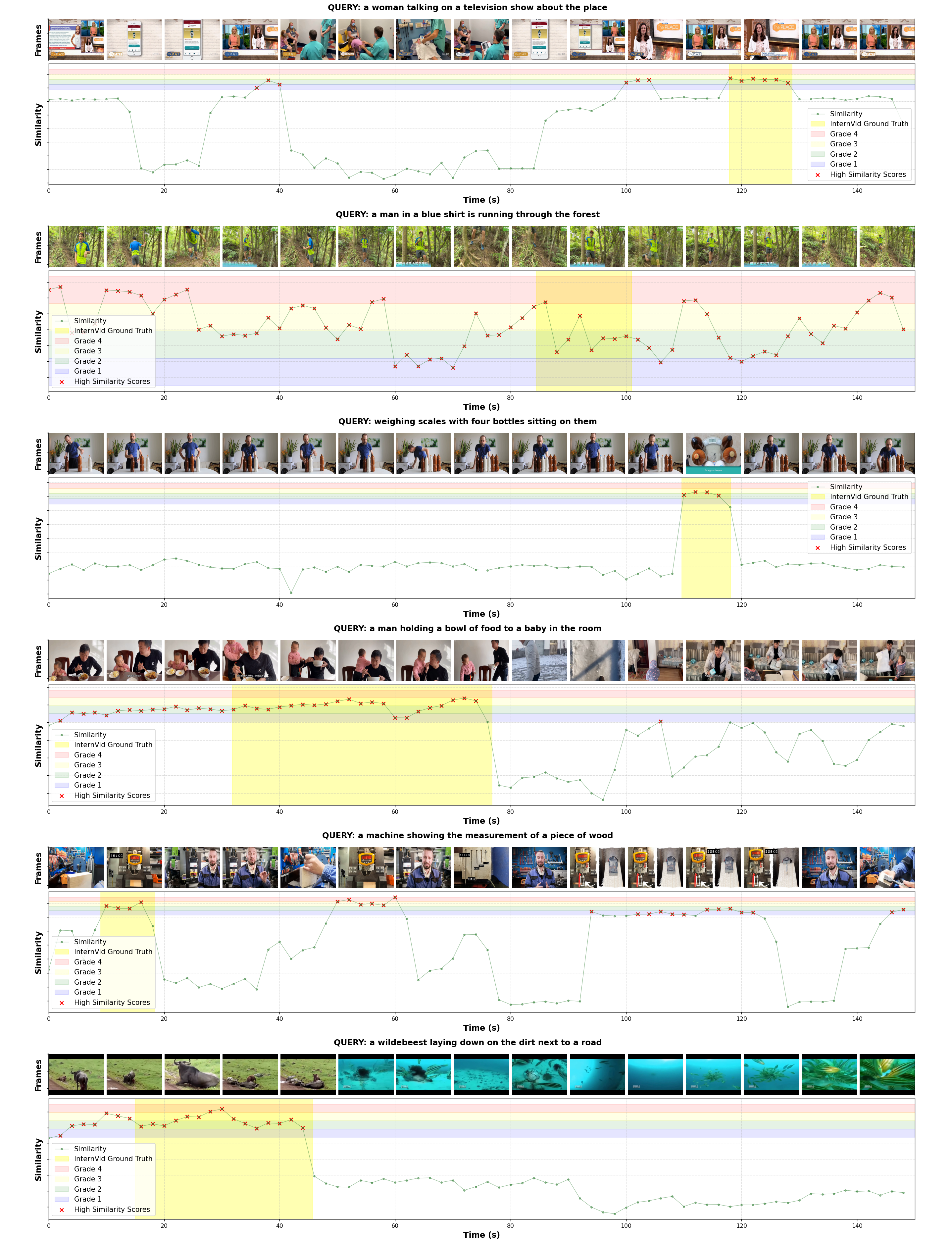}
    \caption{Design of the Preatrain Annotation}
    \label{fig:PretrainAnnoVis}
\end{figure*}

To clearly illustrate how the InterVid-MR pretraining dataset was created, we present key statistics related to the annotation generation process in \cref{fig:PretrainAnnoVis}. Although the annotation process is straightforward, the visualization confirms that the resulting annotations achieve high quality.

\end{document}